\documentclass[conference]{IEEEtran}

\usepackage{acro}
\usepackage[utf8]{inputenc}
\usepackage[ruled,vlined,linesnumbered]{algorithm2e}
\usepackage{amsfonts, amsmath, amssymb}
\usepackage{array}
\usepackage{bm}
\usepackage{cite}
\usepackage{color}
\usepackage{float}
\usepackage{subfig}
\usepackage{hyperref}
\usepackage{url}
\usepackage{multirow}
\usepackage{booktabs}
\usepackage{graphicx}
\usepackage[switch]{lineno}

\title{Underwater Object Classification and Detection: first results and open challenges}

\author{
    André Jesus$^{3*}$, Claudio Zito$^{1*}$, Claudio Tortorici$^{1}$, Eloy Roura$^{1}$, Giulia De Masi$^{1,2}$
    \\
    $^{1}$ Technology Innovation Institute (TII), Abu Dhabi, United Arab Emirates\\
    $^{2}$ College of Engineering, Khalifa University, Abu Dhabi, United Arab Emirates\\
    $^{3}$ Vrije Universiteit Amsterdam, Amsterdam, the Netherlands \\
    a.l.fialhojesus@student.vu.nl, \{\textit{Claudio.Zito, Claudio.Tortorici, Eloy.Roura, Giulia.DeMasi}\}\!\atsign tii.ae\\
    $^*$ Co-first authors who have contributed equally to this work
}

\newcommand\atsign{@}

\newcommand*{\sref}[1]{Section~\ref{#1}}

\newcommand*{\fref}[1]{Figure~\ref{#1}}
\newcommand*{\tref}[1]{Table~\ref{#1}}

\begin{document}
    \maketitle

\begin{abstract}
    This work reviews the problem of object detection in underwater environments. We analyse and quantify the shortcomings of conventional state-of-the-art (SOTA) algorithms in the computer vision community when applied to this challenging environment, as well as providing insights and general guidelines for future research efforts. First, we assessed if pretraining with the conventional ImageNet is beneficial when the object detector needs to be applied to environments that may be characterised by a different feature distribution. We then investigate whether two-stage detectors yields to better performance with respect to single-stage detectors, in terms of accuracy, intersection of union (IoU), floating operation per second (FLOPS), and inference time. Finally, we assessed the generalisation capability of each model to a lower quality dataset to simulate performance on a real scenario, in which harsher conditions ought to be expected. Our experimental results provide evidence that underwater object detection requires searching for ``ad-hoc'' architectures than merely training SOTA architectures on new data, and that pretraining is not beneficial.
\end{abstract}

\section{Introduction}

The ocean is essential for life on our planet and our economy. It works as a global climate control system, and it is an indispensable source of food and energy. 
Yet, it is the least explored habitat due to its harsh conditions that prevent its exploration by conventional means. The world beneath the ocean represents a thriving environment for autonomous robots, and its useful applications varies from exploring deep sea, protecting and preserving its ecosystems, to defence, archaeology, and rescue missions.
Regardless of the application, most underwater robots make use of vision for perceiving the surroundings, and in so object detection plays a critical role in it.

\begin{figure}[t]
    \centering
    \includegraphics[width=.48\textwidth]{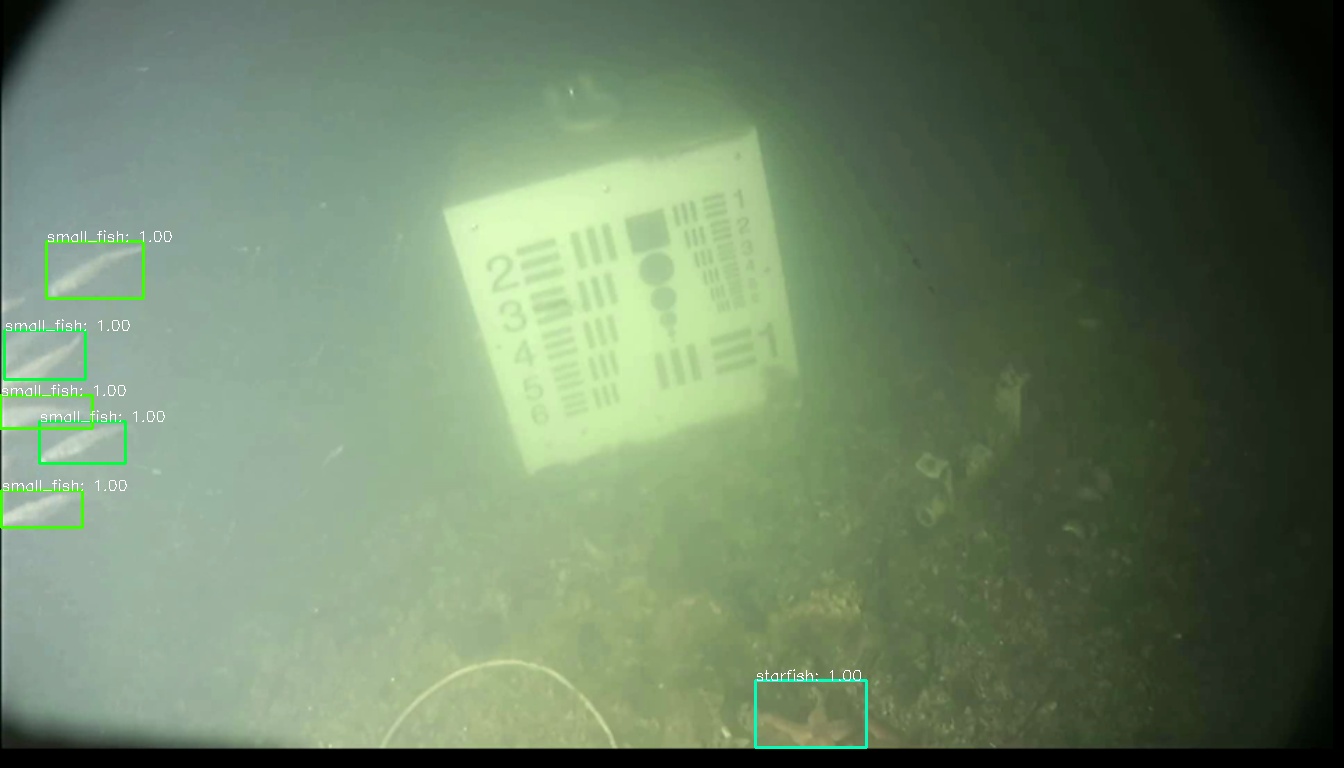}
    \caption{A sample image from \cite{brackish_Pedersen_2019_CVPR_Workshops} showing the harsh conditions met in underwater environments.}
    \label{fig:brackish}
\end{figure}

The data collection and processing for supporting the learning of underwater object detection exposes new challenges. Several unfavourable factors, such as the scattering and absorption of the light by water and the presence of suspended particles, interfere with the image quality. The background acts as a blurry entity that distorts perspective and lessens the contours and colours, rendering techniques such as enhancement and restoration most needed as well as unsatisfactory for underwater vision (\fref{fig:brackish}). Another crucial factor lies in the small averaging sizes of the objects that populate the aquatic environments; current deep learning-based detectors suffer loss in performance on small objects even in conventional settings. Finally, due to low bandwidth communication channels available underwater, the entire process has to be performed within the onboard capabilities, enhancing the need for fast and efficient algorithms. 

The latest trend for generic object detection algorithms mainly relies on Convolutional Neural Network (CNN). However, the SOTA object detectors can be divided into two large categories. On one hand, we have two-stage detectors, such as R-CNN (Region-based CNN) \cite{rcnn2013}, that in a first phase use a Region Proposal Network (RPN) for generating regions of interests on the image. These regions are then fed down the pipeline for the second phase in which classification and bonding-box regression is performed. On the other hand, single-stage detectors, such as Yolo (You Only Look Once) \cite{yolo} and SSD (Single Shot Multibox Detector) \cite{ssd}, treat detection as a regression problem by learning class probabilities and bounding-box coordinates. It is well-known that the former approach achieves higher accuracy in detection at the cost of larger inference times w.r.t. the single-stage counterpart.  
Although efforts have been made to improve efficiency of two-stage detectors by reducing the inference time, the subsequent Fast R-CNN~\cite{girshickICCV15fastrcnn} and Faster R-CNN~\cite{ren2016faster} still do not meet the requirements needed for edge computing devices. 

\begin{figure*}[t]
    \centering
    \includegraphics[width=.95\textwidth]{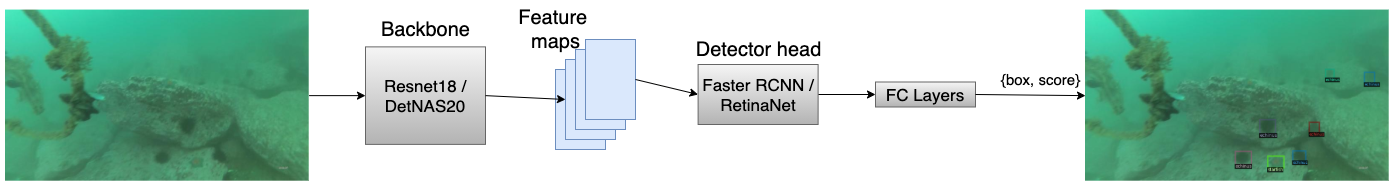}
    \caption{General object detection pipeline. The input layer is followed up by a feature extractor (i.e., backbone), a feature map for ROI localisation, and finally a regression and classification network.}
    \label{fig:high_level_method}
\end{figure*}

Object detection is a more challenging problem than simple object classification since it has to answer to the ``what'' and ``where'' questions for potentially many objects in the same image. Thus, plugging good classifiers in the object detection pipeline not necessarily leads to better detection performance. Figure~\ref{fig:high_level_method} shows the conventional pipeline in which, after the input layer, a backbone network identifies objects within the input image; its output is then fed to the rest of the pipeline, also called \emph{head}. For both single-stage and two-stage detectors, this component is critical and, very recently, identified as a bottleneck--backbones should be trained to support detection and no mere classification--for which a new trend is to select and train better backbones in lieu of outlining better \emph{heads}.    

In this paper, we will compare performance of mainstream algorithms from both classes (single- and two-stage detectors) in order to highlight the strengths and short-comings of conventional object detectors when applied to underwater scenarios. The rest of the paper is structured as follows. We will present previous effort for underwater object recognition and its applications (Section~\ref{sec:relatedwork}), then we will proceed to introduce the chosen datasets, metrics and object detectors, as well as the training and testing procedures we used for each architecture (Section~\ref{sec:matnmet}). Section~\ref{sec:results} will present and discuss our results. We will conclude the paper with our final remarks.

\section{Related Work}\label{sec:relatedwork}

This section presents an overview of how computer vision methods have been applied to assist in underwater monitoring and detection.
There is a considerable amount of research in underwater settings, mostly with biology motivations.  More recently, exploration and research inspired to improve underwater vehicles perception \cite{appli_under} has been at the centre of attention from the scientific community. 
This is due to technological advancements that allow for higher quality footage and sensorial information extraction from the underwater realm.
The two underlying processes in the water-light interactions: scattering and absorption \cite{img_restau_2010} pose several limitations on the reach of existing computer vision methods applied inland.
Filtering and processing approaches such as image enhancement by Fabri et. al. \cite{enh_gans1,enh_gans2} aim at reducing these effects with Generative model approaches.
Another factor in underwater perception development is the performance improvements in machine learning methods over the last decade, in both performance and efficiency which made possible the development of robust approaches without compromising in the real-time constraint. Furthermore, recent advancements in robot grasping and manipulation under uncertainty could be combined with more robust object detectors for exploring novel approach of contact-based tasks underwater in autonomous \cite{bib:zito_w2013,bib:zito_2013,bib:zito_2019} or teleoperated systems \cite{bib:zitoWRRS2019b,10.3389/fnbot.2021.695022,heiwoltICORR2019}. 

The area of object classification became active with recent efforts in the constructions of marine datasets for coral reefs \cite{coral_reef}, fish \cite{fish_wild} and diver-robot underwater interactions in \cite{robot_diver1,robot_diver2}.
Likewise, the applications on Convolutional Neural Networks (CNNs) gave more possibilities to underwater applications as shown by Villon et. al \cite{cnn_and_svm}, while Yang et. al. \cite{yolov3_and_faster} have compared the YOLOv3 and Faster RCNN on a underwater dataset with the results showing a better ability in detecting smaller object by the single-stage architecture. 

Underwater data augmentation via generative modelling approaches are common strategies for enhancement and restoration. 
For instance, Wang et. al. \cite{wang2020_poisson_gan} proposed a Poisson-blending GAN which overcomes some of the common object detection augmentation pitfalls enabling change of the position, number and size of the object classes in a given image.
Chen et. al. \cite{chen2020_rfs_gan_dom} provides an analysis of the image restoration effect in the underwater object detection performance and compares the normal dataset approach to a filter based restoration and a GAN-based restoration of the dataset. Similarly, Yoo et. al. \cite{liu2020_style_transfer} provided a domain generalization for the URPC dataset by applying a style transfer model. This  yields greater performance when compared to training with the normal dataset.

In this work, the object detection methodology takes advantage of the more robust fully-supervised training paradigm in CNNs. Additionally we employ two different underwater dataset designed for object detection. The Brackish dataset~\cite{brackish_Pedersen_2019_CVPR_Workshops} provides us with high quality images comparable to the best enhanced and restored datasets, whilst URPC has been chosen to challenge the detectors with a larger variety of conditions which it is expected to find in real applications at sea.


\section{Materials and Methods}\label{sec:matnmet}

\begin{table*}[t]
\centering
    \begin{tabular}{l|c|c|c|c}
    DetNAS Model & \multicolumn{4}{c}{Architecture configuration (20 blocks in 4-4-8-4)} \\
    \hline
    Faster RCNN w/ ImageNet Pre. & {[}X, 3, 5, 5, & 7, 3, 3, 5, & X, X, X, 3, 7, 7, X, 3, & 7, 3, X, 7{]} \\
    Faster RCNN from scratch & {[}X, 7, 7, X, & X, X, X, 3, & X, X, X, X, 5, 3, 3, 3, & 5, X, X, 7{]} \\
    \hline
    RetinaNet w/ ImageNet Pre. & {[}X, X, 5, 3, & 3, 3, 3, 3, & 3, X, X, X, X, 7, 5, X, & 3, 7, X, X{]} \\
    RetinaNet from scratch & {[}7, 5, 5, 3, & 5, 7, X, 5, & 5, 3, X, X, 5, 7, 7, 3, & 3, 5, 7, 5{]}
    \end{tabular}
    \caption{Best performing architecture obtained in Supernet Search process for Brackish Dataset. The order is from 1st to 20th block. We coded the block types as follows: "shufflenet\_3x3" ($3$), "shufflenet\_5x5" ($5$), "shufflenet\_7x7" ($7$) and "xception\_3x3" ($X$).}
    \label{res:brackish_archs}
\end{table*}

\subsection{Chosen models}

We have selected several SOTA architectures for our evaluations, namely: Faster RCNN (two-stage) \cite{ren2016faster}, RetinaNet (single-stage) \cite{retinanet}, and YOLOv3 (single-stage) \cite{yolo}. The latter is used as a baseline, using results from~\cite{brackish_Pedersen_2019_CVPR_Workshops,yolonano_urpc}, while we analysed the benefits of a customised feature extractor (backbone) for the first two architectures. \fref{fig:high_level_method} shows a graphical representation of the general pipeline. We tested DetNAS20~\cite{detnas} and ResNet18~\cite{he2015deep} backbones on both architectures, hereafter referred to as DetNAS20+FRCNN, DetNAS20+RetNet, ResNet18+FRCNN, and ResNet18+RetNet. Whilst ResNet18 architecture only learns from the data by adapting its weights, DetNAS20 tailors weights and the architecture to the data in a three-stage process: (i) pre-training (optional), (ii) fine-tuning, and (iii) evolutionary search in the architecture space. 

\subsection{Datasets}

Our tests employ two publicly available datasets for underwater object detection: Brackish and URPC2019. Additionally, we employed ImageNet for verifying the benefits of a pre-trained backbone in such tasks.

The Brackish dataset is constituted of 14,518 frames with 25,613 box annotations and six classes: Big fish, Crab, Jellyfish, Shrimp, Small fish and Starfish. The data collection was performed by a fixed station with three mounted cameras on the sea bed. 

The URPC2019\footnote{\url{www.cnurpc.org}} dataset is a collection of footage from the annual Underwater Robotics Competition. It contains 4757 images with 37130 box annotations divided into four classes:  Scallop, Starfish, Echinus and Holothurian. This is a challenging dataset with lower quality images collected by different robots.

\subsection{Data processing}

The data processing applied to the dataset images differs in the DetNAS20 and the Resnet18 models.  On the former, the training preprocessing is characterized by a horizontal flip of the image with a probability $0.5$ and the image resizing to be within the minimum (800) and maximum (1300) size in pixels. 
For the Resnet18 models, the training preprocessing contains a horizontal flip with still $0.5$ probability as in DetNAS20, but the image are adjusted such that the shorter edge is within the minimum (800) and maximum (1333) size. 
The image pixels are normalised both in training and inference.

\subsection{Metrics}

Standard metrics in object detection were used for evaluation and comparison of the models, including accuracy evaluation, such as average Intersection over Union (IoU)--which measures the extension of the overlap between two bounding boxes, i.e. the larger the overlap, the higher is the IoU--and the metrics proposed in COCO~\cite{lin2014microsoft}, i.e., average precision (AP) and average recall (AR); as well as performance evaluation, with Float Operation Per Second (FLOPS) and inference time.

\subsection{Hardware for the experiments}

For the training and testing we have used four Nvidia Titan X with 12 GB of VRAM, except for the stage three in the training of the DetNas20 in which we used two Nvidia RTX2080 TI with 11 GB of VRAM.  

\subsection{Experimental evaluation} 

Our experimental evaluation consists of several comparisons. First, we run the DetNAS supernetwork procedure~\cite{detnas} to search for a tailored architecture. We trained the network in two conditions: (i) pre-trained with ImageNet and (ii) without it. The learning curves in both conditions overlap after very few iterations ($<1$\% of the total number of iterations) revealing that using ImageNet is not beneficial in the underwater domain given the differences in the image distribution. We, hence, proceeded into training, validating and testing our five models on each dataset separately without the use of ImageNet. Next section will discuss our findings and concluding remarks.

\section{Results}\label{sec:results}

\subsection{Pretraining with ImageNet vs training from scratch}\label{sec:pretrained}

\begin{table*}
\centering
\begin{tabular}{ |l||c|c|c|c|c|c|c|c| }
 \hline
  \multicolumn{1}{|c||}{Model} & \multicolumn{4}{c|}{Brackish} & \multicolumn{4}{c|}{URPC} \\
 \hline
  & AP & AP$_{50}$ & FLOPS & Inference & AP & AP$_{50}$ & FLOPS & Inference \\
  \hline
 YOLOv3  & 38.9 & 83.7 & 65.86 & - & - & \textbf{74.6} & 65.86 & - \\
 DetNAS20+FRCNN  & \textbf{78.0} & \textbf{97.0} & 110.077 & 0.127852 / 7.8 & 33.4 & 61.8 & 107.768 & 0.127938 / 7.8  \\
 DetNAS20+RetNet & 77.6 & 96.4 & 125.388 & 0.083831 / 11.9 & 31.8 & 63.8 & 121.890 & 0.088115 / 11.3  \\
 ResNet18+FRCNN & 76.9 & 96.2 & 141.340 & 0.071765 / 13.9 & \textbf{36.2} & 65.6 & 136.643 & 0.077219 / 13.0 \\
 ResNet18+RetNet & 68.8 & 88.3 & 136.652 &0.058148 / 17.2 & 19.7 & 33.6 & 136.387 & 0.058660 / 17.0 \\
 \hline
\end{tabular}
\caption{Summary of our experimental results (without  pre-training) using the COCO metrics AP and AP$_{50}$, in which the values are shown in percentile, and FLOPS (Gi) and inference time (sec / FPS) comparison for both datasets. First row is from~\cite{brackish_Pedersen_2019_CVPR_Workshops, yolonano_urpc}.}
 \label{tab:results_summary}
\end{table*}

\begin{table}
\centering
\begin{tabular}{ |l||c|c|c|c|c|c| }
 \hline
  \multicolumn{1}{|c||}{Model} & \multicolumn{6}{c|}{Brackish (AP$_{50}$)} \\
 \hline
  & Big f. & Crab & Jell. & Shr. & Sma. f. & Star. \\
  \hline
 YOLOv3  & 90.0 & 92.7 & 82.1 & 76.6 & 62.3 & 98.7 \\
 DetNAS20+FRCNN  & 98.0 & 99.0 & 94.8 & 96.8 & 94.5 & 99.0  \\
 DetNAS20+RetNet & 97.5 & 99.0 & 93.0 & 98.3 & 91.7 & 99.0  \\
 ResNet18+FRCNN & 97.0 & 99.0 & 90.1 & 99.9 & 92.6 & 99.0 \\
 ResNet18+RetNet & 94.9 & 97.0 & 82.2 & 85.2 & 71.4 & 99.0 \\
 \hline
\end{tabular}
\caption{Performance over the (six) labelled classes in the Brackish dataset. Respectively, big fish, Crab, jelly fish, shrimp, small fish, and starfish. The performance is measured using the COCO metric AP$_{50}$ and the values are reported in percentile.}
 \label{tab:brackish_per_class}
\end{table}

The first experiment is an ablation study to investigate the impact of pretrained backbones. Each architecture is composed of 4 levels that contains $4+4+8+4=20$ blocks. For each block to search, we have 4 possibilities, either a ShuffleNetv2 block with a kernel size of \{3×3,5×5,7×7\} or replacing the right branch with an Xception block (three repeated separable depthwise 3×3 convolutions). Thus, the search space includes $4^{20}\approx 1.0\text{x}10^{12}$ possible architectures, for which we employ an evolutionary algorithm (EA) with a population of individuals that applies the binary crossover operator and mutation operator at each generation. Given the considerable computational resources and time to perform the training, the search was only performed once per model despite EAs having a stochastic nature.  
The backbone architectures selected in the supernet stage 3 are illustrated in table \ref{res:brackish_archs}, we coded the block types as follows: "shufflenet\_3x3" ($3$), "shufflenet\_5x5" ($5$), "shufflenet\_7x7" ($7$) and "xception\_3x3" ($X$). 

For the Faster RCNN architectures, we can immediately notice that they are characterised by a dominance of small kernel size blocks ($3\text{x}3$), but the not-pretrained architecture contains almost double of the Xception ShufflNetv2 blocks compared to the pretrained counterpart--which are three times deeper than the  $3\text{x}3$ ShufflNetv2 blocks. In contrast, RetinaNet with no pretraining favours large kernel blocks ($5\text{x}5$,$7\text{x}7$) in all levels. 

At the training stage , the learning curves in both conditions (Faster RCNN and RetinaNet) are overlapped after $<1$\% of the total number of iterations, which suggests that using ImageNet in the underwater domain does not help improving the performance given the differences in the image distribution. 

The performance in terms of accuracy is slightly inferior for all the pretrained models, except for the shrimp class in which Faster RCNN scores a $99.0\%$ in the AP$_{50}$ against the $96.8\%$ without ImageNet. FLOPS, IoU, and inference times are consistent in both conditions: YOLOv3 has half the FLOPS complexity than two-stage architectures; DetNAS20 is the second best in FLOPS, but RetNet18 has a faster inference time ($\sim 4$ fps faster); and same distribution for the IoU values (see \fref{fig:iou_bra}). It is worth mentioning that the pretraining seems to have a slightly positive effect on the FLOPS complexity and inference time within the DetNAS models with $109.807$ Gi. and $0.121615/8.2$ inference time (s/fps) versus the $110.77$ Gi. and $0.127852/7.8$ as reported in \tref{tab:results_summary}.

Based on these observations, we discarded the pretrained models and used the models trained from scratch for the next experiments; hereafter renamed DetNAS20+FRCNN for the ``Faster RCNN from scratch'' (\tref{res:brackish_archs} second row) and DetNAS20+RetNet for the ``RetinaNet from scratch'' (\tref{res:brackish_archs} fourth row).  

\subsection{Full comparison on Brackish}

\begin{figure}[t]
\centering
\includegraphics[width=.48\textwidth]{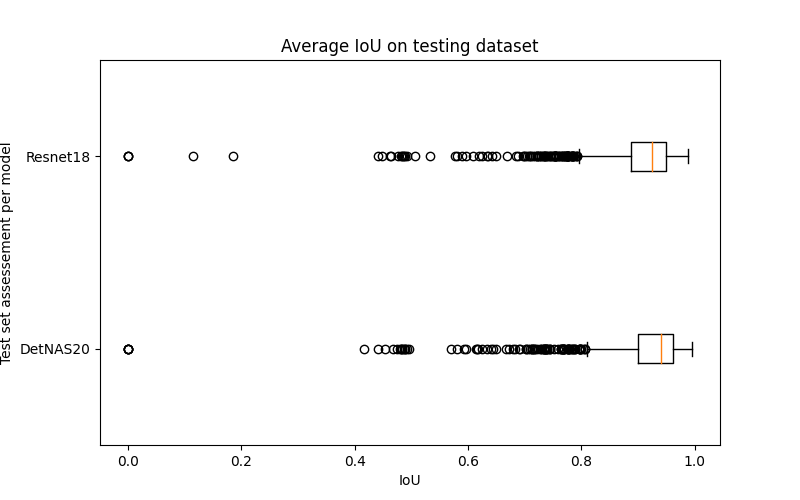}
\caption{Average IoU on the Brackish testing dataset (1239 images). It compares the best performing models: ResNet18+FRCNN (top) and DetNAS20+FRCNN (bottom).}
\label{fig:iou_bra}
\end{figure}

Our main results are summarised in \tref{tab:results_summary}. We employed the two DetNAS20 architectures from \sref{sec:pretrained} trained from scratch on the Brackish dataset. We compared them using two ResNet18 architectures, again trained from scratch on the Brackish dataset, and as a baseline we adopted the results from~\cite{brackish_Pedersen_2019_CVPR_Workshops, yolonano_urpc} for the single-stage architecture YOLOv3.  

On Brackish, DetNas20 demonstrates higher reliability as a backbone than ResNet18. By analysing the performance over the labelled classes in \tref{tab:brackish_per_class}, we cast some clarity on the overall architectures' behaviour. Classes such as \textit{`crab'} and \textit{`starfish'} tends to be easier to detect since the majority of the samples appears on the seafloor, whilst the \textit{`small fish'} and \textit{`jellyfish'} are amongst the harder classes for the models to learn since they can appear in any part of the image with similar frequency and they have relatively small size. The last two rows of \tref{tab:brackish_per_class} clearly show that ResNet18 suffers in identifying useful features even for the easier classes, especially when combined with a RetNet head; $94.9$ and $97.0$ percentile for, respectively, big fish and crab classes; a 2-3 percentile points fewer than the other methods. In combination with Faster R-CNN, the performance improves slightly overall, but not enough to be compared with the DetNas backbone. Interestingly, the accuracy over the shrimp class is at $99.9\%$ for this method, which is the same accuracy we observed from DetNAS20+FRCNN when pretrained with ImageNet. 

On the other hand, YOLOv3 without the support of any backbone scores the lower overall accuracy, as well as per classes, demonstrating once again the benefit of the architectures supported by a backbone in terms of accuracy. In terms of FLOPS, the single-stage network outperforms the other models by half of the complexity; both models with DetNAS20 backbone have fewer FLOPS, but slightly higher inference time, than their counterparts. This is most likely due to the fact that ResNet18 is not ``overly deep'' w.r.t. DetNAS20--starting with an advantage of 2 fewer blocks, and the lack of deeper Xception blocks.   

Finally, \fref{fig:iou_bra} shows the average IoU for both backbones on the Brackish dataset. In this comparison we used the same head (Faster RCNN) since it is the most performing one. Both  methods  show  robust  results  in  their ability to recognize objects in novel images. This is illustrated by the boxplots having  the  quartile  Q1  above  0.85  and  Q3  bellow  0.95  in  a  scale  from  0  to 1.  The  mean  IoU  for  the  1239  test  images  is  slightly  higher  in  DetNAS20  than Resnet18 with 0.9123 and 0.8996, respectively.

\subsection{Full comparison on URPC}

\begin{table}
\centering
\begin{tabular}{ |l||c|c|c|c| }
 \hline
  \multicolumn{1}{|c||}{Model} & \multicolumn{4}{c|}{URPC (AP$_{50}$)} \\
 \hline
  & Holo. & Echi. & Scal. & Star. \\
  \hline
 YOLOv3  & 47.0 & 90.8 & 75.9 & 84.9 \\
 DetNAS20+FRCNN  & 26.8 & 85.5 & 59.9 & 75.2  \\
 DetNAS20+RetNet & 28.4 & 88.1 & 60.8 & 77.9  \\
 ResNet18+FRCNN & 43.3 & 83.0 & 59.8 & 76.4 \\
 ResNet18+RetNet & 4.0 & 56.8 & 22.3 & 51.3 \\
 \hline
\end{tabular}
\caption{Performance over the (four) labelled classes in the URPC dataset. Respectively, holothrurian, echinus, scallop, and starfish. The performance is measured using the COCO metric AP$_{50}$ and the values are reported in percentile.}
 \label{tab:urpc_per_class}
\end{table}

\begin{figure}[t]
\centering
\includegraphics[width=.48\textwidth]{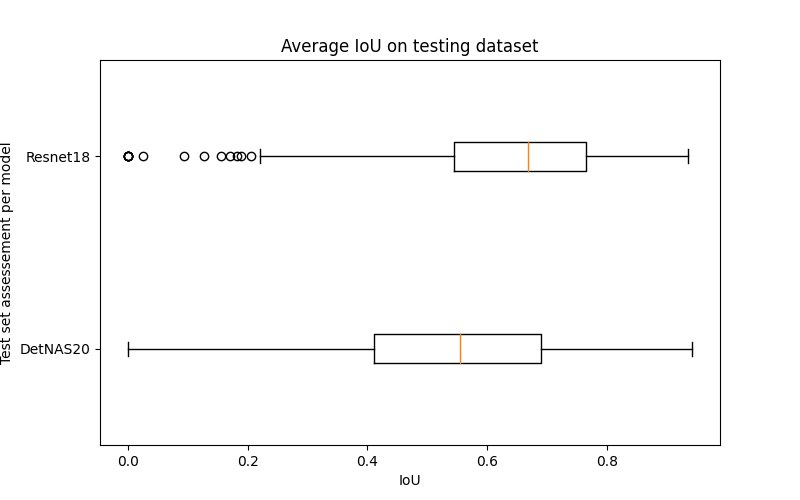}
\caption{Average IoU on the Urpc2019 testing dataset (470 images). It compares the best performing models: ResNet18+FRCNN (top) and DetNAS20+FRCNN (bottom).}
\label{fig:iou_urpc}
\end{figure}

The overall COCO metric results for URPC19 are described in \tref{tab:results_summary} (third section), whilst \tref{tab:urpc_per_class} reports the results by class of objects. For this comparison, we did not retrained our DetNAS supernet on the lower quality URPC images to evaluate the robustness on real scenarios in which it is to expect harsher conditions. YOLOv3 results from \cite{yolonano_urpc} is the highest performing model with a mere $74.6$\% on the AP$_{50}$. In general, the performance of the models is considerably lower in the \textit{'holothurian'} and \textit{'scallop'} classes since they contain a similar color to most backgrounds in the dataset. 
The Resnet18+FRCNN model achieves the closest performance to YOLOv3 even though it only performs the detection at 13 FPS.

The average IoU over the testing URPC19 samples is illustrated in \fref{fig:iou_urpc} and it compares the best performing DetNAS model (Faster RCNN) on the bottom to the best Resnet18 based model (Faster RCNN) on the top part.
The boxplots illustrate their difference in performance with Resnet18 quartile Q1 being similar to the median of the DetNAS20. The images where Resnet18 had a IoU below $0.2$ were considered outliers while in DetNAS20 they are still captured by the boxplot.
It is visible that the mean IoU for the 470 images is higher in the Resnet18 model than in the DetNAS model, $0.64$ and $0.54$ respectively. 
The former is more accurate in the localization of the objects in a scene. 
This is supported by the fact that DetNAS model did not predict any boxes in 5 more images than the Resnet18.

From the Urpc 2019 dataset experiment, the Resnet18 and DetNAS based detectors display similar performance across the different classes; they are however considerably bellow the YOLOv3 performance. 
The reason pointed in \cite{yolov3_and_faster} for this is that the YOLOv3 is better suited to detect very small objects, which characterizes the majority of the dataset samples, due to the multi-scale predictions approached proposed in YOLOv3.

\section{Concluding Remarks}\label{sec:remarks}

This paper presents a review of the problem of object detection for underwater robotics. We compared five different architectures considered as SOTA for object detection on two different publicly available underwater datasets. 

We assessed whether the conventional procedure of pre-training the feature extractor, or backbone, have any effect on the task at hand. In~\cite{he_2019} similar findings show that pre-training does not help generalisation to other tasks, even in conventional ``on the ground'' settings. 
Nevertheless, an ``ad-hoc'' dataset designed to improve edge detection of blurry, colour-faded objects could positively impact the learning. We will investigate such an approach as an extension of this work.

\tref{tab:results_summary} summarises the performance of our chosen models on  Brackish and URPC2019 underwater datasets. DetNas20+FRCNN performs well on Brackish -- on a conventional dataset, such as VOC, this model has an AP$=80\%$ -- but all the models suffer from low accuracy when faced with the URPC. Generalise over a larger variety of harsh conditions has a profound impact on performance which suggests that a larger search space should be considered when training the DetNAS supernet. Future research directions should focus on enhancement and reconstruction techniques as pre-processing steps. Another interesting area to explore is to combine the DetNAS supernet with semi-supervised object detection so that the architectures learn from partially labelled sources.    
\section*{Acknowledgment}
C.T. and G.DM. would like to acknowledge the support from the project funded by the Technology Innovation Institute “Heterogeneous Swarm of Underwater Autonomous Vehicles”  developed with Khalifa University (Project no. TII/ARRC/2047/2020).


\bibliographystyle{IEEEtran}
\bibliography{references}

\begin{thebibliography}{10}
\providecommand{\url}[1]{#1}
\csname url@samestyle\endcsname
\providecommand{\newblock}{\relax}
\providecommand{\bibinfo}[2]{#2}
\providecommand{\BIBentrySTDinterwordspacing}{\spaceskip=0pt\relax}
\providecommand{\BIBentryALTinterwordstretchfactor}{4}
\providecommand{\BIBentryALTinterwordspacing}{\spaceskip=\fontdimen2\font plus
\BIBentryALTinterwordstretchfactor\fontdimen3\font minus
  \fontdimen4\font\relax}
\providecommand{\BIBforeignlanguage}[2]{{%
\expandafter\ifx\csname l@#1\endcsname\relax
\typeout{** WARNING: IEEEtran.bst: No hyphenation pattern has been}%
\typeout{** loaded for the language `#1'. Using the pattern for}%
\typeout{** the default language instead.}%
\else
\language=\csname l@#1\endcsname
\fi
#2}}
\providecommand{\BIBdecl}{\relax}
\BIBdecl

\bibitem{brackish_Pedersen_2019_CVPR_Workshops}
M.~Pedersen, J.~Bruslund~Haurum, R.~Gade, and T.~B. Moeslund, ``Detection of
  marine animals in a new underwater dataset with varying visibility,'' in
  \emph{Proceedings of the IEEE/CVF Conference on Computer Vision and Pattern
  Recognition (CVPR) Workshops}, June 2019.

\bibitem{rcnn2013}
R.~Girshick, J.~Donahue, T.~Darrell, and J.~Malik, ``Rich feature hierarchies
  for accurate object detection and semantic segmentation,'' \emph{Proceedings
  of the IEEE Computer Society Conference on Computer Vision and Pattern
  Recognition}, 11 2013.

\bibitem{yolo}
J.~Redmon, S.~Divvala, R.~Girshick, and A.~Farhadi, ``You only look once:
  Unified, real-time object detection,'' 06 2016, pp. 779--788.

\bibitem{ssd}
W.~Liu, D.~Anguelov, D.~Erhan, C.~Szegedy, S.~Reed, C.-Y. Fu, and A.~Berg,
  ``Ssd: Single shot multibox detector,'' vol. 9905, 10 2016, pp. 21--37.

\bibitem{girshickICCV15fastrcnn}
R.~Girshick, ``Fast r-cnn,'' in \emph{International Conference on Computer
  Vision ({ICCV})}, 2015.

\bibitem{ren2016faster}
S.~Ren, K.~He, R.~Girshick, and J.~Sun, ``Faster r-cnn: Towards real-time
  object detection with region proposal networks,'' \emph{IEEE Transactions on
  Pattern Analysis and Machine Intelligence}, vol.~39, 06 2015.

\bibitem{appli_under}
N.~Gracias, R.~Garcia, R.~Campos, N.~Hurtos, R.~Prados, A.~Shihavuddin,
  T.~Nicosevici, A.~Elibol, and J.~Escartin, \emph{Application Challenges of
  Underwater Vision: Land, Sea \& Air}, 02 2017, pp. 133--160.

\bibitem{img_restau_2010}
R.~Schettini and S.~Corchs, ``Underwater image processing: State of the art of
  restoration and image enhancement methods,'' \emph{EURASIP Journal on
  Advances in Signal Processing}, vol. 2010, 12 2010.

\bibitem{enh_gans1}
C.~{Fabbri}, M.~J. {Islam}, and J.~{Sattar}, ``Enhancing underwater imagery
  using generative adversarial networks,'' in \emph{2018 IEEE International
  Conference on Robotics and Automation (ICRA)}, 2018, pp. 7159--7165.

\bibitem{enh_gans2}
M.~J. Islam, Y.~Xia, and J.~Sattar, ``Fast underwater image enhancement for
  improved visual perception,'' 2020.

\bibitem{bib:zito_w2013}
C.~Zito, M.~Kopicki, R.~Stolkin, C.~Borst, F.~Schmidt, M.~A. Roa, and J.~Wyatt,
  ``Sequential re-planning for dextrous grasping under object-pose
  uncertainty,'' in \emph{Workshop on Manipulation with Uncertain Models,
  Robotics: Science and Systems ({RSS})}, 2013.

\bibitem{bib:zito_2013}
------, ``Sequential trajectory re-planning with tactile information gain for
  dextrous grasping under object-pose uncertainty,'' in \emph{{IEEE} Proc.
  Intelligent Robots and Systems ({IROS})}, 2013.

\bibitem{bib:zito_2019}
C.~Zito, V.~Ortienzi, M.~Adjigble, M.~S. Kopicki, R.~Stolkin, and J.~L. Wyatt.,
  ``Hypothesis-based belief planning for dexterous grasping,'' \emph{CoRR arXiv
  preprint, arXiv:1903.05517 [cs.RO] (cs.AI)}, 2019.

\bibitem{bib:zitoWRRS2019b}
C.~Zito, M.~Adjigble, B.~Denoun, L.~Jamone, M.~Hansard, and R.~Stolkin,
  ``Metrics and benchmarks for remote shared controllers in industrial
  applications,'' in \emph{Proc. of the Workshop on Task-Informed Grasping
  (TIG-II): From Perception to Physical Interaction, Robotics: Science and
  Systems ({RSS})}, 2019.

\bibitem{10.3389/fnbot.2021.695022}
\BIBentryALTinterwordspacing
S.~Veselic, C.~Zito, and D.~Farina, ``Human-robot interaction with robust
  prediction of movement intention surpasses manual control,'' \emph{Frontiers
  in Neurorobotics}, vol.~15, p. 125, 2021. [Online]. Available:
  \url{https://www.frontiersin.org/article/10.3389/fnbot.2021.695022}
\BIBentrySTDinterwordspacing

\bibitem{heiwoltICORR2019}
K.~Heiwolt, C.~Zito, M.~Nowak, C.~Castellini, and R.~Stolkin, ``Automatic
  detection of myocontrol failures based upon situational context
  information,'' in \emph{Proceeding of {IEEE/RAS-EMBS} International
  Conference on Rehabilitation Robotics ({ICORR})}, 2019.

\bibitem{coral_reef}
O.~{Beijbom}, P.~J. {Edmunds}, D.~I. {Kline}, B.~G. {Mitchell}, and
  D.~{Kriegman}, ``Automated annotation of coral reef survey images,'' in
  \emph{2012 IEEE Conference on Computer Vision and Pattern Recognition}, 2012,
  pp. 1170--1177.

\bibitem{fish_wild}
\BIBentryALTinterwordspacing
P.~Zhuang, Y.~Wang, and Y.~Qiao, ``Wildfish: A large benchmark for fish
  recognition in the wild,'' in \emph{Proceedings of the 26th ACM International
  Conference on Multimedia}, ser. MM '18.\hskip 1em plus 0.5em minus
  0.4em\relax New York, NY, USA: Association for Computing Machinery, 2018, p.
  1301–1309. [Online]. Available:
  \url{https://doi.org/10.1145/3240508.3240616}
\BIBentrySTDinterwordspacing

\bibitem{robot_diver1}
\BIBentryALTinterwordspacing
A.~Gomez~Chavez, A.~Ranieri, D.~Chiarella, E.~Zereik, A.~Babić, and A.~Birk,
  ``Caddy underwater stereo-vision dataset for human–robot interaction (hri)
  in the context of diver activities,'' \emph{Journal of Marine Science and
  Engineering}, vol.~7, no.~1, 2019. [Online]. Available:
  \url{https://www.mdpi.com/2077-1312/7/1/16}
\BIBentrySTDinterwordspacing

\bibitem{robot_diver2}
M.~J. Islam, M.~Fulton, and J.~Sattar, ``Towards a generic diver-following
  algorithm: Balancing robustness and efficiency in deep visual detection,''
  2018.

\bibitem{cnn_and_svm}
S.~Villon, M.~Chaumont, G.~Subsol, S.~Villéger, T.~Claverie, and D.~Mouillot,
  ``Coral reef fish detection and recognition in underwater videos by
  supervised machine learning: Comparison between deep learning and hog+svm
  methods,'' vol. 10016, 10 2016, pp. 160--171.

\bibitem{yolov3_and_faster}
H.~Yang, P.~Liu, Y.~Hu, and J.~Fu, ``Research on underwater object recognition
  based on yolov3,'' \emph{Microsystem Technologies}, 01 2020.

\bibitem{wang2020_poisson_gan}
Z.~Wang, C.~Liu, S.~Wang, T.~Tang, Y.~Tao, C.~Yang, H.~Li, X.~Liu, and X.~Fan,
  ``Udd: An underwater open-sea farm object detection dataset for underwater
  robot picking,'' 2020.

\bibitem{chen2020_rfs_gan_dom}
X.~Chen, Y.~Lu, Z.~Wu, J.~Yu, and L.~Wen, ``Reveal of domain effect: How visual
  restoration contributes to object detection in aquatic scenes,'' 2020.

\bibitem{liu2020_style_transfer}
H.~Liu, P.~Song, and R.~Ding, ``Wqt and dg-yolo: towards domain generalization
  in underwater object detection,'' 2020.

\bibitem{retinanet}
T.-Y. Lin, P.~Goyal, R.~Girshick, K.~He, and P.~Dollar, ``Focal loss for dense
  object detection,'' 10 2017, pp. 2999--3007.

\bibitem{yolonano_urpc}
L.~Wang, X.~Ye, H.~Xing, Z.~Wang, and P.~Li, ``Yolo nano underwater: A fast and
  compact object detector for embedded device,'' in \emph{Global Oceans 2020:
  Singapore – U.S. Gulf Coast}, 2020, pp. 1--4.

\bibitem{detnas}
\BIBentryALTinterwordspacing
Y.~Chen, T.~Yang, X.~Zhang, G.~Meng, C.~Pan, and J.~Sun, ``Detnas: Neural
  architecture search on object detection,'' \emph{CoRR}, vol. abs/1903.10979,
  2019. [Online]. Available: \url{http://arxiv.org/abs/1903.10979}
\BIBentrySTDinterwordspacing

\bibitem{he2015deep}
K.~He, X.~Zhang, S.~Ren, and J.~Sun, ``Deep residual learning for image
  recognition,'' in \emph{Proceedings of the IEEE conference on computer vision
  and pattern recognition}, 2016, pp. 770--778.

\bibitem{lin2014microsoft}
T.-Y. Lin, M.~Maire, S.~Belongie, J.~Hays, P.~Perona, D.~Ramanan,
  P.~Doll{\'a}r, and C.~L. Zitnick, ``Microsoft coco: Common objects in
  context,'' in \emph{European conference on computer vision}.\hskip 1em plus
  0.5em minus 0.4em\relax Springer, 2014, pp. 740--755.

\bibitem{he_2019}
K.~He, R.~Girshick, and P.~Dollar, ``Rethinking imagenet pre-training,'' in
  \emph{2019 IEEE/CVF International Conference on Computer Vision (ICCV)},
  2019, pp. 4917--4926.

\end{thebibliography}
\end{document}